\documentclass[conference]{IEEEtran}
\IEEEoverridecommandlockouts
\usepackage{cite}
\usepackage{amsmath,amssymb,amsfonts}
\usepackage{graphicx}
\usepackage{textcomp}
\usepackage{algorithm}
\usepackage{algpseudocode}
\usepackage{xcolor}
\usepackage{amssymb}            
\usepackage{mathtools}          
\usepackage{mathrsfs}           
\mathtoolsset{showonlyrefs}     
\usepackage{graphicx}           
\usepackage{subcaption}         
\usepackage[space]{grffile}     
\usepackage{url}                
\usepackage{amsmath}
\usepackage{tikz}
\usepackage{bm}
\usepackage{float}
\usepackage{booktabs} 
\usetikzlibrary{arrows.meta, positioning, fit, backgrounds, shadows}
\usepackage{pgfplots}
\pgfplotsset{compat=1.18}  
\usepgfplotslibrary{fillbetween}

\usepackage{caption}       
\usepackage{dblfloatfix}   
\usepackage{placeins}  
\usepackage[hidelinks]{hyperref}

\usetikzlibrary{
  positioning,   
  fit,           
  shapes,        
  arrows.meta,   
  backgrounds,   
  calc           
}
\newcommand{\znorm}[1]{\mathcal{Z}(#1)}

\def\BibTeX{{\rm B\kern-.05em{\sc i\kern-.025em b}\kern-.08em
    T\kern-.1667em\lower.7ex\hbox{E}\kern-.125emX}}
\begin{document}

\title{Adaptive Correlation-Weighted Intrinsic Rewards for Reinforcement Learning\\
}

\author{\IEEEauthorblockN{Viet Bac Nguyen}
\IEEEauthorblockA{\textit{Institute for Artificial Intelligence}\\
\textit{VNU-University of Engineering and Technology}\\
Hanoi, Vietnam\\
bacnv.hv@uet.edu.vn}
\and
\IEEEauthorblockN{Phuong Thai Nguyen}
\IEEEauthorblockA{\textit{Institute for Artificial Intelligence}\\
\textit{VNU-University of Engineering and Technology}\\
Hanoi, Vietnam\\
thainp@vnu.edu.vn}
}

\maketitle

\begin{abstract}
We propose ACWI (Adaptive Correlation-Weighted Intrinsic), an adaptive intrinsic reward scaling framework designed to dynamically balance intrinsic and extrinsic rewards for improved exploration in sparse reward reinforcement learning. Unlike conventional approaches that rely on manually tuned scalar coefficients, which often result in unstable or suboptimal performance across tasks, ACWI learns a state dependent scaling coefficient online.
Specifically, ACWI introduces a lightweight Beta Network that predicts the intrinsic reward weight directly from the agent state through an encoder based architecture. The scaling mechanism is optimized using a correlation based objective that encourages alignment between the weighted intrinsic rewards and discounted future extrinsic returns. This formulation enables task adaptive exploration incentives while preserving computational efficiency and training stability.
We evaluate ACWI on a suite of sparse reward environments in MiniGrid. Experimental results demonstrate that ACWI consistently improves sample efficiency and learning stability compared to fixed intrinsic reward baselines, achieving superior performance with minimal computational overhead.
\end{abstract}


\section{Introduction}

\begin{figure}[t]
    \centering
    \includegraphics[width=\linewidth]{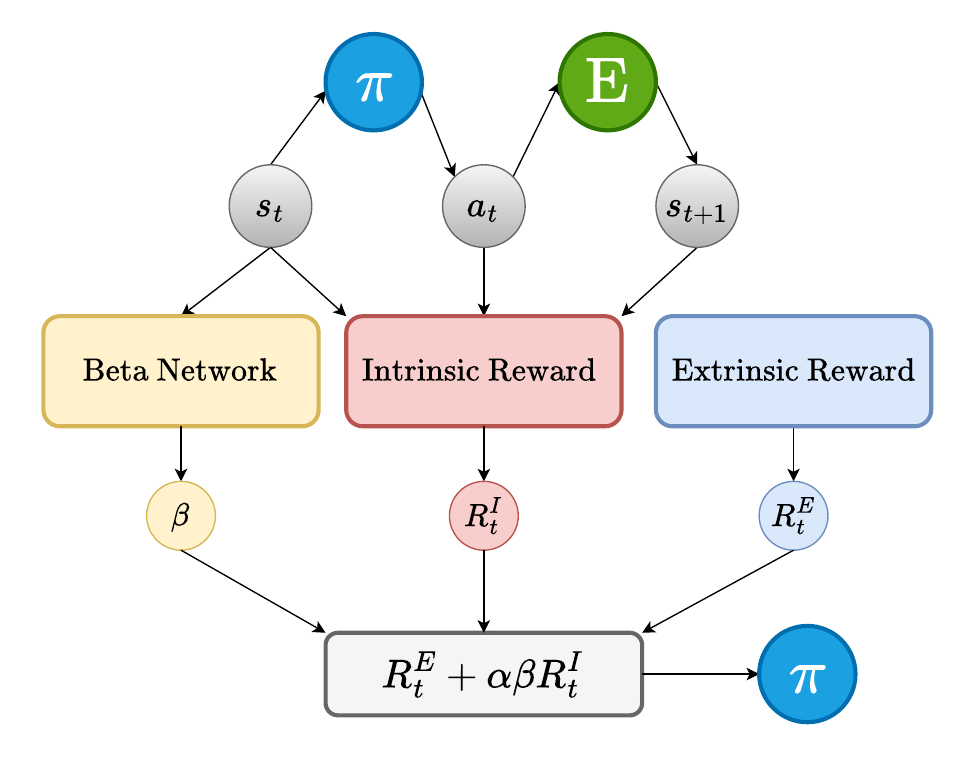}
    \caption{\textbf{Overview of ACWI.} Given state $s_t$, action $a_t$,
    and next state $s_{t+1}$ in environment $E$, the Beta Network produces a state-dependent
    scaling factor $\beta(s_t)$. The extrinsic reward $R^{E}_{t}$ and
    intrinsic reward $R^{I}_{t}$ are computed from the transition
    $(s_t, a_t, s_{t+1})$. The combined signal
    $R^{E}_{t} + \alpha\beta(s_t) R^{I}_{t}$ is used to update the policy
    $\pi$, where $\alpha$ is a global coefficient controlling the overall
    magnitude of the intrinsic reward relative to the extrinsic reward.}
    \label{fig:overview}
\end{figure}

Reinforcement learning (RL) has achieved remarkable success in settings
with dense and informative reward signals \cite{mnih2015,silver2016}.
However, learning in sparse-reward or long-horizon environments remains
a fundamental challenge, as the agent receives little feedback to
distinguish productive behaviors from random ones
\cite{sutton2018,francois2018}.
Classical exploration strategies such as $\varepsilon$-greedy or
Boltzmann exploration address this by assigning non-zero probability to
all actions \cite{sutton2018}, yet they treat all unexplored actions
equally and become increasingly inefficient as the observation space
grows in dimensionality.

To provide more structured exploration incentives, a large body of work
proposes augmenting the extrinsic reward with an intrinsic motivation
signal \cite{bellemare2016,ostrovski2017,tang2017,pathak2017,burda2018_rnd,raileanu2020}.
Count-based methods define intrinsic rewards inversely proportional to
state visitation frequency, encouraging the agent to revisit
infrequently seen states \cite{bellemare2016,tang2017}.
Building on this, curiosity-driven methods such as Intrinsic Curiosity Module (ICM) \cite{pathak2017}, Random Network Distillation (RND) \cite{burda2018_rnd}, and Rewarding Impact-Driven Exploration (RIDE) \cite{raileanu2020} provide richer intrinsic signals by leveraging prediction errors or representation discrepancies to quantify novelty and learning progress. Although differing in their specific formulations, these approaches share a common principle: they derive intrinsic rewards from prediction errors or discrepancies in learned representations, thereby encouraging exploration of states that are difficult to predict or insufficiently modeled.
Importantly, these curiosity-driven methods exhibit a natural self-regulating behavior: as the agent repeatedly visits the same states, prediction errors decrease, representations stabilize, and the associated intrinsic rewards gradually diminish. This property prevents persistent over-exploration of already familiar regions and promotes a progressive shift toward less-explored parts of the state space.

Despite this self-regulating property, a fundamental limitation
persists in how intrinsic rewards are combined with extrinsic ones.
The standard formulation mixes the two signals through a fixed scalar
coefficient \cite{pathak2017,burda2018_rnd}, selected
by manual hyperparameter search.
While intrinsic reward \emph{magnitudes} do decay over time for
familiar states, the fixed coefficient cannot distinguish between
states where continued exploration is genuinely useful for downstream
task performance and states where it is not.
Two states may receive similar intrinsic bonuses yet differ greatly in
their strategic value: exploration departing from one may consistently
lead to high extrinsic returns, while exploration from the other may
yield little beyond novelty.
A uniform scaling factor is blind to this distinction and applies the
same weight to intrinsic signals regardless of how well they align with
task-relevant progress.

Several works have attempted to address this limitation.
Extrinsic-Intrinsic Policy Optimization (EIPO) \cite{chen2022eipo}
formulates the trade-off as a constrained policy optimization problem,
automatically tuning the importance of intrinsic rewards so that
exploration is suppressed when it is no longer needed.
Automatic Intrinsic Reward Shaping (AIRS) \cite{yuan2023airs} treats
the selection among multiple intrinsic reward functions as a multi-armed
bandit problem, switching between them based on estimated task return.
While these methods represent meaningful progress, they adapt the
weighting at the level of training stages or reward function types,
without accounting for variation across individual states.
An agent that has thoroughly explored one region of the state space but
remains uncertain in another would benefit from stronger intrinsic
incentives specifically in the latter, a fine-grained adaptation that
global mechanisms cannot provide.

In this paper, we address this gap with \textbf{ACWI} (Adaptive
Correlation-Weighted Intrinsic), a method that learns a state-dependent
scaling function to modulate intrinsic rewards during training.
As illustrated in Figure~\ref{fig:overview}, ACWI introduces a
lightweight Beta Network that operates alongside any intrinsic reward
module, predicting a per-state multiplier that is combined with the
intrinsic signal before it is added to the extrinsic reward.
Our main contributions are as follows:

\begin{itemize}
    \item First, we formulate intrinsic-reward modulation as learning a
    state-dependent multiplier $\beta(s_t)$, predicted by a lightweight
    \emph{Beta Network}.
    The augmented reward takes the form
    $\bar{r}_t = R^{\text{E}}_t + \alpha\,\beta(s_t)\,R^{\text{I}}_t$,
    where $\alpha > 0$ is a global intrinsic-strength coefficient.
    Unlike fixed-coefficient baselines, $\beta(s_t)$ allows the agent
    to amplify exploration incentives in states where exploration tends
    to yield task-relevant progress and suppress them elsewhere.

    \item Second, we introduce a correlation-based training objective
    for the Beta Network that aligns the weighted intrinsic signal
    $\beta(s_t)\,R^{\text{I}}_t$ with the discounted future extrinsic
    return, providing a direct and stable training signal without
    requiring additional policy optimization procedures.

    \item Finally, we integrate ACWI with ICM and evaluate it under
    Proximal Policy Optimization (PPO) \cite{schulman2017ppo} on
    sparse-reward benchmarks.
    Empirical results demonstrate improved sample efficiency and more
    stable learning dynamics compared to fixed-weight baselines.
\end{itemize}
\section{Related Work}
A large body of work augments extrinsic rewards with intrinsic motivation signals to encourage exploration in sparse-reward environments \cite{oudeyer2007,schmidhuber2010}. Count-based methods define intrinsic rewards inversely proportional to state visitation frequency, rewarding the agent for visiting novel states \cite{bellemare2016,tang2017,ostrovski2017}. \cite{strehl2008} provided theoretical guarantees for count-based bonuses in tabular settings, while \cite{bellemare2016} extended the idea to high-dimensional spaces using pseudo-count derived from a density model. Curiosity-driven methods instead ground intrinsic rewards in forward prediction error. The Intrinsic Curiosity Module (ICM) proposed by \cite{pathak2017} learns a task-relevant state representation through an inverse-forward dynamics model and uses the forward prediction error as an exploration bonus. \cite{burda2018_rnd} proposed Random Network Distillation (RND), which measures novelty as the discrepancy between a fixed random target network and a learned predictor, avoiding explicit dynamics modeling while retaining the self-diminishing property of prediction-error rewards. \cite{raileanu2020} proposed RIDE, which defines intrinsic rewards based on the impact of actions on the environment's learned state representation, encouraging the agent to seek transitions that are both novel and causally influenced by its decisions. While these methods differ in their formulation, they share a common structure: as the agent gains experience, intrinsic signals naturally diminish for familiar states, providing a form of implicit exploration scheduling.

Most intrinsic motivation methods combine intrinsic and extrinsic rewards via a fixed scalar coefficient selected through manual tuning \cite{pathak2017,burda2018_rnd}. Empirical evidence suggests that performance is highly sensitive to this choice, with different environments and training phases favoring substantially different weightings \cite{badia2020_agent57,taiga2020}. Common remedies include normalizing, clipping, or applying variance stabilization to intrinsic rewards \cite{burda2018_rnd,badia2020_ngu}, which improve numerical stability but do not account for whether intrinsic signals are informative for achieving downstream task objectives. Several works have sought more principled alternatives.
The theoretical foundation for augmenting rewards with auxiliary signals is established by potential-based reward shaping \cite{ng1999}, which guarantees policy invariance under additive shaping derived from a potential function. In practice, however, most intrinsic reward methods do not enforce consistency between the shaping signal and long-term extrinsic return. As a result, improperly weighted intrinsic rewards can distract the agent from the primary task objective and degrade performance relative to optimizing extrinsic rewards alone \cite{chen2022eipo}. To derive more meaningful exploration signals that mitigate aimless curiosity, from an information-theoretic perspective, \cite{houthooft2016} proposed to maximize the information gain about the agent's belief of environment dynamics, arguing that exploration should be driven by quantities relevant to future prediction. \cite{kim2019} defined intrinsic rewards by maximizing mutual information between state-action representations and dynamics under a linear model. These approaches implicitly link exploration incentives to predictive relevance but do not directly optimize the alignment between intrinsic signals and future extrinsic outcomes. Auxiliary task learning offers a related direction, using additional prediction objectives to improve representation quality and sample efficiency \cite{jaderberg2017,mirowski2017}; more recent work considers adapting auxiliary loss weights based on gradient alignment or learning progress \cite{zahavy2020_stac}, though such methods are not specifically designed to regulate exploration bonuses.

Closest to our approach are methods that adapt the influence of intrinsic motivation in a data-driven manner. \cite{badia2020_ngu} introduced a family of policies trained with different discount factors and intrinsic reward strengths, combined through a mixture to balance exploration and exploitation at the population level. \cite{badia2020_agent57} extended this idea by learning to select among the mixture components using a meta-controller, allowing the agent to adaptively shift between exploratory and exploitative behaviors over the course of training. However, both approaches modulate reward weighting at the level of subpolicies rather than individual states, leaving open the question of how intrinsic incentives should vary as a function of the agent's current situation within the state space.

\section{Background}

\subsection{Markov Decision Processes}

We consider reinforcement learning in the framework of episodic Markov Decision Processes (MDP) \cite{sutton2018}, defined by the tuple
\(
\mathcal{M} = \langle\mathcal{S}, \mathcal{A}, \mathcal{P}, r, \gamma, \rho_0\rangle,
\)
where $\mathcal{S}$ is the state space, $\mathcal{A}$ is the action space, 
$\mathcal{P}: \mathcal{S} \times \mathcal{A} \times \mathcal{S} \rightarrow [0,1]$ denotes the transition kernel,
$\mathcal{R}: \mathcal{S} \times \mathcal{A} \rightarrow \mathbb{R}$ is the reward function,
$\gamma \in [0,1)$ is the discount factor,
and $\rho_0$ is the initial state distribution \cite{sutton2018}.
A (stochastic) policy is defined as a mapping
\(
\pi : \mathcal{S} \rightarrow \mathcal{P}(\mathcal{A}),
\)
where $\mathcal{P}(\mathcal{A})$ denotes the space of probability distributions over the action space.
Equivalently, the policy can be represented as a conditional distribution
\(
\pi(a \mid s),
\)
which specifies the probability of selecting action $a$ given state $s$.

The transition dynamics are characterized by the conditional distribution
\(
\mathcal{P}(s' \mid s, a),
\)
which specifies the probability (or probability density, in the continuous case) of transitioning to state $s'$ after executing action $a$ in state $s$.
We write transitions compactly as
\(
s_{t+1} \sim \mathcal{P}(\cdot \mid s_t, a_t),
\)
where the symbol $\sim$ indicates that the random variable $s_{t+1}$ is sampled according to the corresponding probability distribution.

At each timestep $t$, the agent observes the current state $s_t$, selects an action
\(
a_t \sim \pi_\theta(\cdot \mid s_t),
\)
according to a parameterized stochastic policy $\pi_\theta$, transitions
to the next state $s_{t+1}$, and receives an extrinsic reward
$R_t^{\mathrm{E}} = \mathcal{R}(s_t, a_t)$, which we denote by \( r_t \) for brevity.
A trajectory of length $T$ is defined as
\(
\tau = (s_0, a_0, r_0, s_1, a_1, r_1, \dots, s_T),
\)
where $s_0 \sim \rho_0$, $a_t \sim \pi_\theta(\cdot \mid s_t)$, and
$s_{t+1} \sim \mathcal{P}(\cdot \mid s_t, a_t)$.
The objective is to learn policy parameters $\theta$ that maximize the expected discounted return
\[
J_{\mathrm{E}}(\theta)
=
\mathbb{E}_{\tau \sim (\pi_\theta, \mathcal{P})}\!\left[
\sum_{t=0}^{T-1} \gamma^t R_t^{\mathrm{E}}
\right],
\]
where the expectation is taken over trajectories induced by the policy $\pi_\theta$ and the environment dynamics $\mathcal{P}$.

\subsection{Intrinsic Motivation}

In sparse-reward environments, standard policy optimization methods often struggle due to insufficient exploration. Intrinsic motivation addresses this challenge by augmenting learning with intrinsic rewards based on novelty, surprise, or uncertainty \cite{pathak2017,burda2018_rnd}. 
At each timestep $t$, the agent may receive both extrinsic reward $R_t^{\mathrm{E}}$ and intrinsic reward $R_t^{\mathrm{I}}$.

A common formulation combines these signals into a shaped reward
\(
\bar{r}_t = R_t^{\mathrm{E}} + \beta \, R_t^{\mathrm{I}},
\)
where $\beta \ge 0$ controls the relative influence of intrinsic motivation \cite{pathak2017,burda2018_rnd}. 
The agent is then optimized with respect to the objective
\[
J_{\mathrm{E+I}}(\theta)
=
\mathbb{E}_{\tau \sim (\pi_\theta, P)}\!\left[
\sum_{t=0}^{T-1} \gamma^t
\big(
R_t^{\mathrm{E}} + \beta \, R_t^{\mathrm{I}}
\big)
\right]
\]

Although this formulation is widely adopted, the coefficient $\beta$ is typically treated as a fixed hyperparameter and tuned manually \cite{taiga2020}. 
Its value can strongly affect learning dynamics, and the appropriate balance between intrinsic and extrinsic rewards may vary across tasks, training stages, or states. This observation motivates approaches that seek to adapt the strength of intrinsic motivation automatically rather than fixing it a priori.
\section{Method}

We propose a reinforcement learning framework augmented with intrinsic 
motivation and a correlation-driven adaptive reward scaling mechanism. 
Although agnostic to both the choice of RL algorithm and intrinsic reward 
module, we instantiate our framework on PPO \cite{schulman2017ppo} for its 
compatibility with online learning, stable training dynamics, and strong 
empirical performance, and adopt ICM \cite{pathak2017} for its 
computational efficiency and effectiveness in sparse-reward settings. 
Central to our approach is a learnable state-dependent scaling factor 
optimized via a correlation objective that aligns intrinsic bonuses with 
extrinsic performance, enabling adaptive balancing of exploration and 
exploitation across the state space.

\subsection{Intrinsic Curiosity Module}
\label{sec:icm}

To encourage systematic exploration in sparse-reward environments, we employ an Intrinsic Curiosity Module (ICM) \cite{pathak2017} that provides auxiliary rewards based on the agent's prediction error in a learned feature space. 
Following the original formulation, raw states are first mapped into a latent representation via a feature encoder 
$\varphi: \mathcal{S} \to \mathbb{R}^d$ parameterized by $\eta$.
The module then consists of two complementary components trained in a self-supervised manner: 
a forward dynamics model 
$f^{F}_{\eta}: \mathbb{R}^d \times \mathcal{A} \to \mathbb{R}^d$ 
and an inverse dynamics model 
$f^{I}_{\eta}: \mathbb{R}^d \times \mathbb{R}^d \to \mathcal{A}$.
Given a state transition tuple $(s_t, a_t, s_{t+1}) \in \mathcal{S} \times \mathcal{A} \times \mathcal{S}$, 
let $\varphi_t = \varphi(s_t)$ and $\varphi_{t+1} = \varphi(s_{t+1})$ denote the corresponding feature representations.
The forward model predicts the next feature representation from the current feature and action, 
while the inverse model infers the intervening action from consecutive feature embeddings.
The ICM parameters $\eta$ are optimized by minimizing a weighted combination of forward prediction loss (mean squared error) and inverse prediction loss (cross-entropy):

\begin{equation}
\begin{split}
\mathcal{L}_{\mathrm{ICM}}(\eta) 
&= \alpha_{F} \cdot \mathbb{E}_{(s_t,a_t,s_{t+1}) \sim \mathcal{D}}\!\left[
\frac{1}{2}\left\| f^{F}_{\eta}(\varphi_t, a_t) - \varphi_{t+1} \right\|_2^2
\right] \\
&\quad + \alpha_{I} \cdot \mathbb{E}_{(s_t,a_t,s_{t+1}) \sim \mathcal{D}}\!\left[
-\log p_{\eta}(a_t \mid \varphi_t, \varphi_{t+1})
\right],
\end{split}
\label{eq:icm_loss}
\end{equation}
where $\mathcal{D}$ denotes the on-policy trajectory batch collected under the current policy, 
and $\alpha_F, \alpha_I \in \mathbb{R}_+$ are scalar coefficients balancing the two objectives.

The raw intrinsic reward at timestep $t$ is defined as the squared $\ell_2$ forward prediction error in feature space:
\begin{equation}
I_t \coloneqq 
\frac{1}{2}\left\| 
f^{F}_{\eta}(\varphi_t, a_t) - \varphi_{t+1} 
\right\|_2^2 
\in \mathbb{R}_+.
\label{eq:intrinsic_raw}
\end{equation}

To mitigate scale variations across different environment regions and stabilize gradient flow, 
we apply batch-wise standardization followed by rectification. 
Let $\mu_I$ and $\sigma_I$ denote the empirical mean and standard deviation of $\{ I_t \}_{t=1}^N$ 
within the current minibatch of size $N$. 
The preprocessed intrinsic reward is:

\begin{equation}
I_t^{+} \coloneqq
\max\left(0,\; \frac{I_t - \mu_I}{\sigma_I + \varepsilon}\right) 
\in \mathbb{R}_+,
\label{eq:intrinsic_norm}
\end{equation}

where $\varepsilon > 0$ is a numerical stability constant. 
The rectification suppresses transitions whose normalized prediction error falls below the batch mean, 
retaining only relatively surprising transitions.

\subsection{State-Dependent Intrinsic Reward Scaling}
\label{sec:adaptive_beta}

A central challenge in intrinsic motivation is determining how much to weigh curiosity-driven exploration relative to exploitation of known rewarding behaviors. We address this by introducing a learnable, \emph{state-dependent} scaling factor $\beta_\psi: \mathcal{S} \to \mathbb{R}_{+}$ that adaptively modulates the contribution of intrinsic rewards based on local state properties. The augmented reward signal used throughout the policy learning pipeline is:
\begin{equation}
\bar{r}_t \coloneqq R_t^{\mathrm{E}} + \alpha \cdot \beta_\psi(s_t) \cdot I_t^{+},
\label{eq:combined_reward}
\end{equation}
where $R_t^{\mathrm{E}} \in \mathbb{R}$ is the extrinsic reward provided by the environment, $I_t^{+}$ is the normalized intrinsic reward from Eq.~\eqref{eq:intrinsic_norm}, $\alpha \in \mathbb{R}_{+}$ is a fixed global hyperparameter controlling the overall magnitude of intrinsic motivation, and $\beta_\psi(s_t)$ provides fine-grained, state-specific modulation. All downstream computations, including Generalized Advantage Estimation (GAE) \cite{schulman2015gae}, value function targets, and policy gradient estimates, operate on $\bar{r}_t$.

\paragraph{Parameterization}
We parameterize $\beta_\psi(s)$ as a small neural network mapping states to strictly positive scalars. A $L$-layer fully-connected encoder with Tanh activations produces a state embedding, which a two-layer MLP head maps to a scalar log-factor clamped to $[\log\beta_{\min}, \log\beta_{\max}]$, guaranteeing $\beta_\psi(s) \in [\beta_{\min}, \beta_{\max}]$.

\paragraph{Correlation Objective}
Rather than employing computationally expensive second-order meta-learning algorithms \cite{finn2017}, we design a lightweight first-order objective that directly encourages alignment between scaled intrinsic rewards and subsequent task performance. The core principle is simple: intrinsic bonuses should be amplified in states that lead to high extrinsic returns and suppressed elsewhere.

Formally, let $\mathcal{G}_t^{\mathrm{E}} \coloneqq \sum_{k=t}^{T} \gamma^{k-t} R_k^{\mathrm{E}}$ denote the discounted extrinsic return from timestep $t$ onward, where $\gamma \in [0,1)$ is the discount factor \cite{sutton2018}. We seek to maximize the correlation between the scaled intrinsic signal $\mathcal{I}_t \coloneqq \beta_\psi(s_t) \cdot I_t^{+}$ and the extrinsic return $\mathcal{G}_t^{\mathrm{E}}$.

To eliminate scale dependencies and ensure numerical stability, we standardize both quantities within each minibatch $\mathcal{B} = \{(s_i, \mathcal{I}_i, \mathcal{G}_i^{\mathrm{E}})\}_{i=1}^N$:
\begin{equation}
\hat{\mathcal{I}}_i 
\coloneqq 
\frac{\mathcal{I}_i - \mu_\mathcal{I}}{\sigma_\mathcal{I}},
\qquad
\hat{\mathcal{G}}_i 
\coloneqq 
\frac{\mathcal{G}_i^{\mathrm{E}} - \mu_\mathcal{G}}{\sigma_{\mathcal{G}}},
\label{eq:standardize}
\end{equation}

where the empirical minibatch statistics are defined as
\begin{align}
\mu_\mathcal{I} 
&\coloneqq 
\mathbb{E}_{\mathcal{B}}[\mathcal{I}], 
&
\sigma_\mathcal{I}^2 
&\coloneqq 
\mathbb{E}_{\mathcal{B}}\!\left[(\mathcal{I}-\mu_\mathcal{I})^2\right] + \varepsilon, \\
\mu_\mathcal{G} 
&\coloneqq 
\mathbb{E}_{\mathcal{B}}[\mathcal{G}^{\mathrm{E}}], 
&
\sigma_\mathcal{G}^2 
&\coloneqq 
\mathbb{E}_{\mathcal{B}}\!\left[(\mathcal{G}^{\mathrm{E}}-\mu_\mathcal{G})^2\right] + \varepsilon,
\label{eq:batch_stats}
\end{align}
with $\mathbb{E}_{\mathcal{B}}[\cdot]$ denoting the empirical expectation over the current minibatch $\mathcal{B}$ and $\varepsilon > 0$ a small constant for numerical stability.

The correlation objective is then defined as
\begin{equation}
L_{\mathrm{corr}}(\psi)
\coloneqq
-
\mathbb{E}_{\mathcal{B}}
\!\left[
\hat{\mathcal{I}}\,\hat{\mathcal{G}}
\right]
=
-
\frac{
\mathbb{E}_{\mathcal{B}}
\!\left[
(\mathcal{I}-\mu_\mathcal{I})
(\mathcal{G}^{\mathrm{E}}-\mu_\mathcal{G})
\right]
}{
\sigma_\mathcal{I}\,\sigma_\mathcal{G}
}.
\label{eq:meta_corr}
\end{equation}
Minimizing $L_{\mathrm{corr}}$ encourages $\beta_\psi$ to upweight intrinsic bonuses in states that precede high extrinsic returns, effectively steering exploration toward task-relevant regions of the state space. 
To mitigate correlation noise and prevent $\beta_\psi$ from collapsing to extreme values, we introduce an $\ell_2$ regularization in log-space:
\begin{equation}
L_{\mathrm{reg}}(\psi)
\coloneqq
\mathbb{E}_{\mathcal{B}}
\!\left[
\big(\log \beta_\psi(s) - \log \beta_0 \big)^2
\right].
\label{eq:meta_reg}
\end{equation}
The complete objective is
\begin{equation}
L_{\beta}(\psi)
\coloneqq
L_{\mathrm{corr}}(\psi)
+
\lambda_{\mathrm{reg}}\, L_{\mathrm{reg}}(\psi),
\label{eq:meta_total}
\end{equation}
where $\lambda_{\mathrm{reg}} \in \mathbb{R}_{+}$ controls the trade-off between alignment and stability.

\paragraph{Optimization}
The Beta network parameters $\psi$ are updated via gradient descent on $L_{\beta}$ once per training iteration, immediately prior to the $K$ epochs of PPO policy updates:
\begin{equation}
\psi \leftarrow \psi - \alpha_{\psi} \cdot \nabla_{\psi} L_{\beta}(\psi),
\label{eq:psi_update}
\end{equation}
where $\alpha_{\psi}$ is the learning rate and gradients are clipped to a maximum norm for stability \cite{pascanu2013}. Critically, during this update, the policy parameters $\theta$ are held fixed (detached from the computational graph), ensuring that the Beta network optimization does not introduce second-order dependencies into the PPO update. This single update occurs before PPO processes the batch using the augmented rewards from Eq.~\eqref{eq:combined_reward}.
\subsection{Policy Optimization with PPO}

The policy is optimized using PPO \cite{schulman2017ppo} with the augmented reward signal $\bar{r}_t$ from Eq.~\eqref{eq:combined_reward}. We first compute advantages using Generalized Advantage Estimation (GAE) \cite{schulman2015gae}. For the combined reward signal $\bar{r}_t$, the temporal difference residual is
\begin{equation}
    \delta_t = \bar{r}_t + \gamma V_\phi(s_{t+1}) - V_\phi(s_t)
\end{equation}
and the advantage estimate is
\begin{equation}
    \hat{A}_t = \sum_{l=0}^{\infty} (\gamma \lambda)^l \delta_{t+l}.
\end{equation}
Advantages are normalized to zero mean and unit variance before optimization \cite{sutton2018}.
The policy is then updated using a clipped surrogate objective:
\begin{equation}
L_{\text{PPO}}(\theta)
=
\mathbb{E}_t \left[
    \min \left(
        \rho_t(\theta)\,\hat{A}_t,
        \text{clip}(\rho_t(\theta), 1 - \epsilon, 1 + \epsilon)\,\hat{A}_t
    \right)
\right]
\end{equation}

where
\(
\rho_t(\theta)
=
\pi_\theta(a_t \mid s_t) / \pi_{\theta_{\text{old}}}(a_t \mid s_t)
\)
is the importance sampling ratio.
The full objective is defined as
\begin{equation}
L(\theta, \phi)
=
- L_{\text{PPO}}(\theta)
+ c_v \, \mathbb{E}_t \left[(V_\phi(s_t) - {\mathcal{G}^{E}_t})^2\right]
- c_e \, \mathbb{E}_t [H_t],
\end{equation}
where $H_t = H(\pi_\theta(\cdot \mid s_t))$ denotes the entropy of the policy at time step $t$.
The complete training procedure is summarized in Algorithm~\ref{alg:adaptive_beta}.

\begin{algorithm}
\caption{PPO with Correlation-Based Adaptive Intrinsic Scaling}
\label{alg:adaptive_beta}
\begin{algorithmic}[1]
\State \textbf{Input:}
\State $\pi_\theta$, $V_\phi$: Policy and value networks
\State $\mathcal{M}_\eta = \{\varphi_\eta, f^{F}_\eta, f^{I}_\eta\}$: ICM module
\State $\beta_\psi$: Beta network 
\State $\alpha$: Intrinsic strength
\State $K, E$: PPO and ICM epochs
\State \textbf{Define:}
\State $\mathcal{Z}(\cdot)$: z-score normalization with zero mean, unit variance over batch
\While{not converged}
    \State $\mathcal{D} := \{(s_t, a_t, r_t, s_{t+1}, d_t)\}_{t=0}^{T-1}$ using $\pi_{\theta}$
    
    \For{$e = 1$ to $E$}
        \State Update $\eta$ by minimizing $\mathcal{L}_{\text{ICM}}$ on minibatches 
        \State from $\mathcal{D}$
    \EndFor
    
    \State $I^{+}_t$ := intrinsic reward compute according \eqref{eq:intrinsic_norm}
    
    \State Compute return $\mathcal{G}^{\text{E}}_t$ for each $t$
    \State Compute $\mathcal{I}_t := \beta_\psi(s_t) \cdot I^{+}_t$ for each $t$
    \State Compute $\hat{\mathcal{I}}_t := \znorm{\mathcal{I}_t}$ and $\hat{\mathcal{G}}_t := \znorm{\mathcal{G}_t^{\text{E}}}$
    \State Compute correlation $(\hat{\mathcal{I}}_t, \hat{\mathcal{G}}_t)$ and update $\psi$ by 
    \State minimizing $L_{\beta}(\psi)$ according \eqref{eq:meta_total}
    
    \State Compute $\bar{r}_t := r_t + \alpha \cdot \beta_\psi(s_t) \cdot I^{+}_t$ for each $t$
    \State Compute GAE advantages $\hat{A}_t$ using $\bar{r}_t$
    
    \For{$k = 1$ to $K$}
        \State Update $\pi_\theta$ and $V_\phi$ by PPO using $\hat{A}_t$
    \EndFor
\EndWhile
\end{algorithmic}
\end{algorithm}

\begin{table}[t]
\centering
\caption{Training hyperparameters for PPO with ICM and ACWI. Fixed intrinsic coefficients for PPO+ICM baselines are $\beta \in \{0.1,0.2,0.5,1,2\}$.}
\label{tab:hyperparams}
\small
\begin{tabular}{@{}ll@{}}
\toprule
\textbf{Parameter} & \textbf{Value} \\
\midrule
\multicolumn{2}{@{}l}{\textit{PPO configuration}} \\
Discount factor $\gamma$ & 0.99 \\
GAE parameter $\lambda$ & 0.95 \\
PPO epochs $K$ & 4 \\
Clip range $\epsilon$ & 0.2 \\
Actor learning rate & $3 \times 10^{-4}$ \\
Critic learning rate & $3 \times 10^{-4}$ \\
\midrule
\multicolumn{2}{@{}l}{\textit{ICM configuration}} \\
ICM learning rate & $10^{-3}$ \\
ICM batch size & 64 \\
Fixed intrinsic coefficients & $\beta \in \{0.1,0.2,0.5,1,2\}$ \\
Global intrinsic strength $\alpha$ & 0.001 \\
\midrule
\multicolumn{2}{@{}l}{\textit{Beta network}} \\
Learning rate & $5 \times 10^{-4}$ \\
Encoding size & 256 \\
Encoder depth $L$ & 2 \\
Output bounds & $[0.1,\,2.0]$ \\
\midrule
\multicolumn{2}{@{}l}{\textit{Correlation objective}} \\
Regularization weight $\lambda_{\mathrm{reg}}$ & $10^{-3}$ \\
Reference value $\beta_0$ & 1.0 \\
Gradient clipping & 1.0 \\
Weight decay & $10^{-6}$ \\
\bottomrule
\end{tabular}
\end{table}

\section{Experiments}

\begin{figure}[H]
    \centering
    \includegraphics[width=\columnwidth]{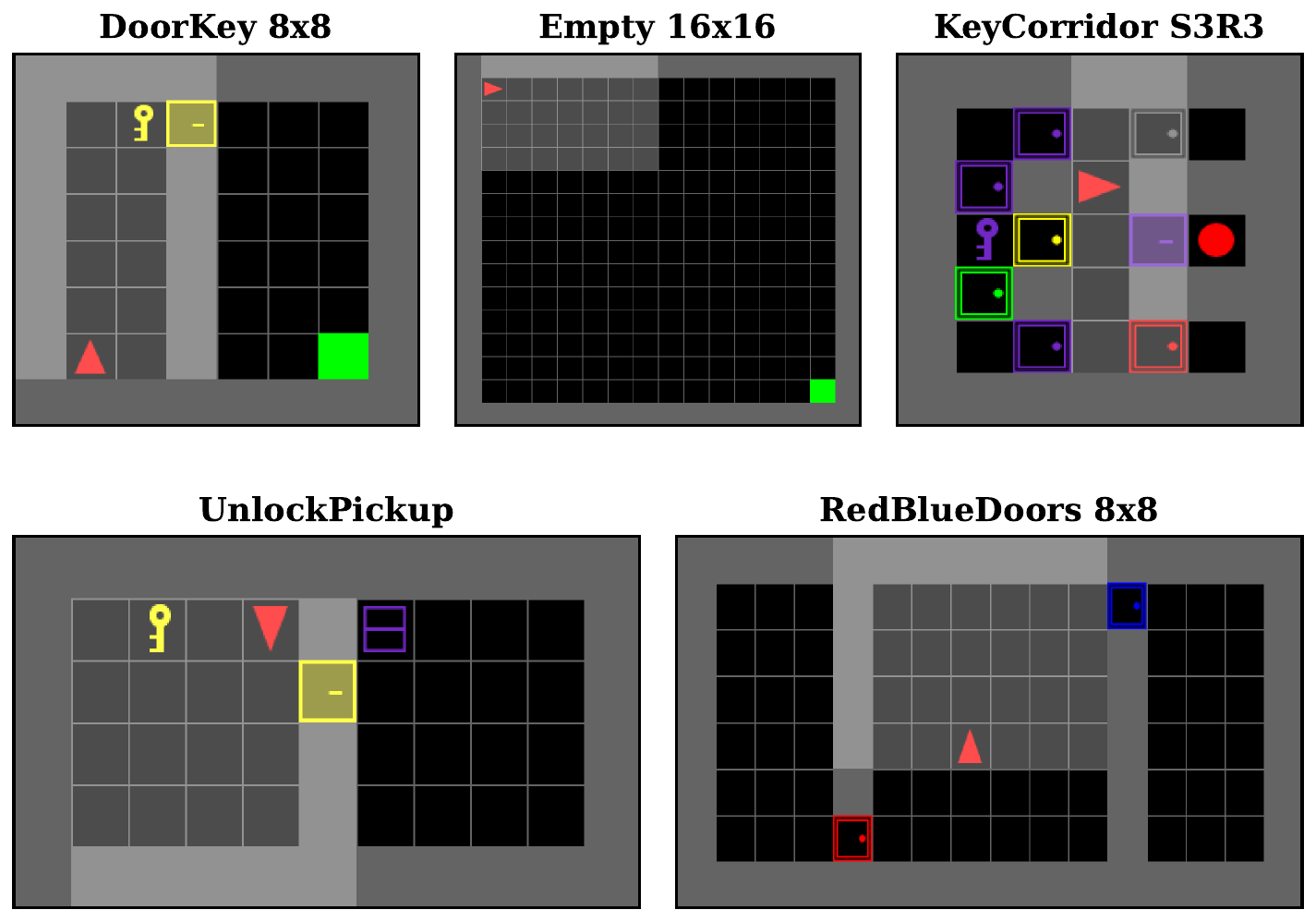}
    \caption{Screenshots of the five MiniGrid environments used in our experiments \cite{chevalier2018minigrid}. From left to right: DoorKey-8x8, Empty-16x16, KeyCorridorS3R3, UnlockPickup, and RedBlueDoors-8x8. Each environment presents increasing complexity in terms of required exploration depth and compositional reasoning.}
    \label{fig:minigrid_envs}
\end{figure}

We evaluate ACWI on five MiniGrid environments~\cite{chevalier2018minigrid} that present distinct challenges for adaptive intrinsic reward scaling (Fig.~\ref{fig:minigrid_envs}). These environments are specifically chosen to test whether ACWI can automatically balance exploration and exploitation across different reward structures without manual tuning:

\begin{itemize}
    \item \textbf{DoorKey-8x8}: The agent must pick up a key and use it to unlock a door before reaching the goal. This environment tests ACWI's ability to \textit{recognize compositional sub-goals}: intrinsic rewards should be upweighted during key search but downweighted once the door is unlocked, as the path to goal becomes clear.
    
    \item \textbf{Empty-16x16}: A large empty room where the agent starts at a random position and must reach the goal square. This environment represents an \textit{extreme sparse-reward regime} with essentially no intermediate extrinsic signals. It tests whether ACWI degrades gracefully when the correlation objective receives minimal informative gradients.
    
    \item \textbf{RedBlueDoors-8x8}: The agent must open the red door first, then the blue door, in a fixed sequence. This tests ACWI's ability to \textit{distinguish task-relevant from task-irrelevant exploration}: the agent should learn to suppress intrinsic rewards for the incorrect door order while maintaining exploration for the correct sequence.
    
    \item \textbf{UnlockPickup}: The agent must unlock a door and pick up a box in another room. Similar to DoorKey but with an additional manipulation requirement, this environment tests ACWI's \textit{adaptation across hierarchical sub-tasks} with minimal intermediate feedback between stages.
    
    \item \textbf{KeyCorridorS3R3}: A multi-room environment with three locked doors and three keys hidden in different rooms. The agent must explore to find keys and navigate through corridors. This is the most challenging environment, testing ACWI's ability to \textit{sustain long-horizon exploration} (often $>$200 steps) while progressively shifting toward exploitation as sub-goals are discovered.
\end{itemize}

All environments share common characteristics that make them suitable for evaluating intrinsic motivation methods: (1) \textbf{sparse terminal rewards} with only success/failure signals, (2) \textbf{egocentric partial observations} that require memory and spatial reasoning, (3) \textbf{step penalties} that discourage random exploration, and (4) \textbf{stochastic initial states} sampled from 5 random seeds to prevent memorization of fixed trajectories and ensure robust generalization.

We compare ACWI against: (i) PPO receiving only extrinsic rewards (lower bound), and (ii) PPO+ICM with fixed intrinsic coefficients $\beta \in \{0.1,0.2,0.5,1,2\}$ (values chosen from preliminary tuning as the most effective range). To ensure fairness, all methods share identical network architectures and training hyperparameters. We fix the global intrinsic strength $\alpha=0.001$ and allow the learned $\beta_\psi(s)$ to be clipped to $[0.1,2.0]$; all other hyperparameters remain unchanged across runs (Table~\ref{tab:hyperparams}). We report episode returns and intrinsic reward magnitudes averaged over 5 random seeds. Comparisons evaluate robustness to fixed-$\beta$ selection, sample efficiency, and stability under varying degrees of extrinsic signal informativeness.

\section{Results}

\begin{figure*}[!h]
    \centering
    \includegraphics[width=\linewidth,height=1.0\textheight,keepaspectratio]{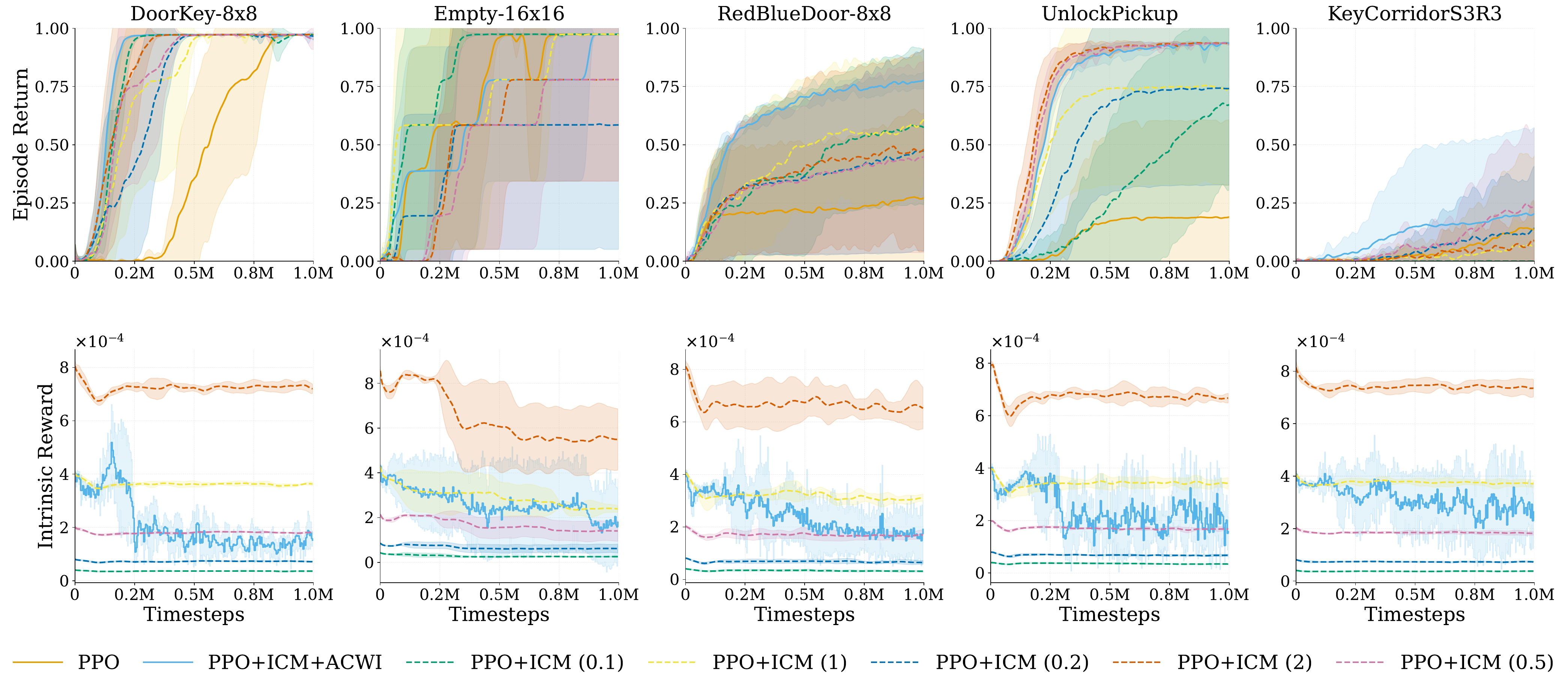}
    \caption{Episode returns over training steps across five MiniGrid environments \cite{chevalier2018minigrid}. Comparison between the adaptive beta mechanism (ACWI), ICM with fixed intrinsic scaling coefficients $\beta \in \{0.1,0.2,0.5,1,2\}$, and the PPO baseline.}
    \label{fig:extrinsic_rewards_comparison}
\end{figure*}

Figure~\ref{fig:extrinsic_rewards_comparison} present learning curves comparing our adaptive beta approach against ICM with fixed intrinsic scaling coefficients $\beta \in \{0.1,0.2,0.5,1,2\}$ and baseline PPO without intrinsic motivation. Solid lines represent mean episode returns averaged over 5 random seeds; shaded regions indicate standard deviation.

Several consistent trends emerge from Figure~\ref{fig:extrinsic_rewards_comparison}, highlighting both the strengths and limitations of correlation-based adaptive intrinsic reward scaling. Across environments with sparse but informative extrinsic rewards, ACWI consistently achieves strong or best performance, demonstrating improved robustness to intrinsic reward scaling compared to fixed-$\beta$ ICM. While fixed intrinsic coefficients can yield competitive results when carefully tuned for a specific task, their performance varies substantially across environments and random seeds. In contrast, ACWI adapts the intrinsic contribution online, resulting in more stable learning dynamics without environment-specific hyperparameter tuning.

The benefits of adaptivity are most pronounced in tasks such as \textit{DoorKey-8x8}, \textit{RedBlueDoors-8x8}, \textit{UnlockPickup}, and \textit{KeyCorridorS3R3}, where extrinsic rewards are sparse but not entirely absent. In these settings, ACWI exhibits faster early learning and reduced variance across seeds, indicating improved sample efficiency and more reliable exploration. Notably, fixed-$\beta$ methods are sensitive to over- or under-scaling: overly aggressive intrinsic rewards can dominate the learning signal and destabilize training, while conservative scaling often leads to insufficient exploration. The adaptive mechanism mitigates both failure modes by modulating intrinsic rewards based on their predictive relationship to future extrinsic returns.

In environments where extrinsic rewards eventually become informative, the intrinsic reward under ACWI gradually decreases as training progresses, reflecting a natural shift from exploration to exploitation. This decay indicates that ACWI learns to suppress intrinsic noise once the policy has identified task-relevant behaviors. In contrast, fixed-$\beta$ intrinsic rewards remain approximately constant throughout training, potentially injecting unnecessary variance even after policy convergence.

A notable exception is the \textit{Empty-16x16} environment, which represents an extreme sparse-reward regime where extrinsic rewards are almost always zero until the agent reaches the goal. In this case, ACWI does not yield a clear advantage over fixed intrinsic scaling. Because the discounted extrinsic return is typically zero for most trajectories, the correlation signal used to train the Beta network collapses, preventing meaningful adaptation. This effect is reflected in the intrinsic reward curves, which remain relatively smooth but exhibit high variance across random seeds. Once a non-zero extrinsic reward is finally observed, the adaptive scaling rapidly stabilizes, and the learned $\beta$ converges with reduced variability. However, prior to this event, ACWI effectively behaves like a fixed intrinsic coefficient.

Overall, these results suggest that adaptive intrinsic reward scaling primarily improves learning stability and sample efficiency rather than targeting peak asymptotic performance. ACWI is most effective in environments where extrinsic rewards, while sparse, provide occasional informative signals that can guide the correlation-based correlation objective. In extremely sparse settings with near-zero extrinsic feedback, the adaptive mechanism degenerates gracefully to fixed intrinsic scaling, highlighting both the practical robustness and the fundamental limitations of correlation-driven adaptation.

\section{Analysis of results}

\begin{figure*}[!h]
    \centering
    \includegraphics[width=\linewidth,height=1.0\textheight,keepaspectratio]{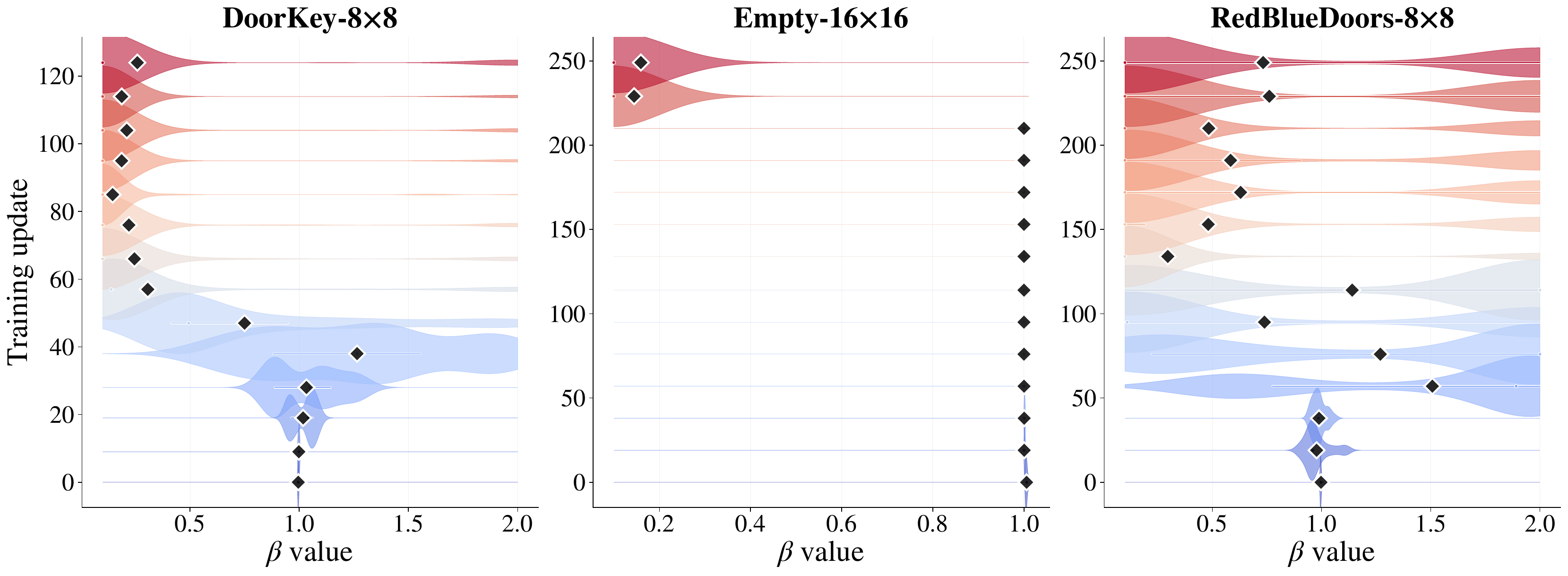}
    \caption{Evolution of the learned state-dependent adaptive $\beta$ distributions during training across three MiniGrid environments. 
    \textit{DoorKey-8x8} and \textit{RedBlueDoors-8x8} progressively develop structured and multimodal distributions, 
    whereas \textit{Empty-16x16} maintains a relatively narrow distribution throughout training, 
    indicating limited adaptation under extreme reward sparsity.}
    \label{fig:beta_distribution_stages}
\end{figure*}

\begin{figure*}[!h]
    \centering
    \includegraphics[width=\linewidth,height=1.0\textheight,keepaspectratio]{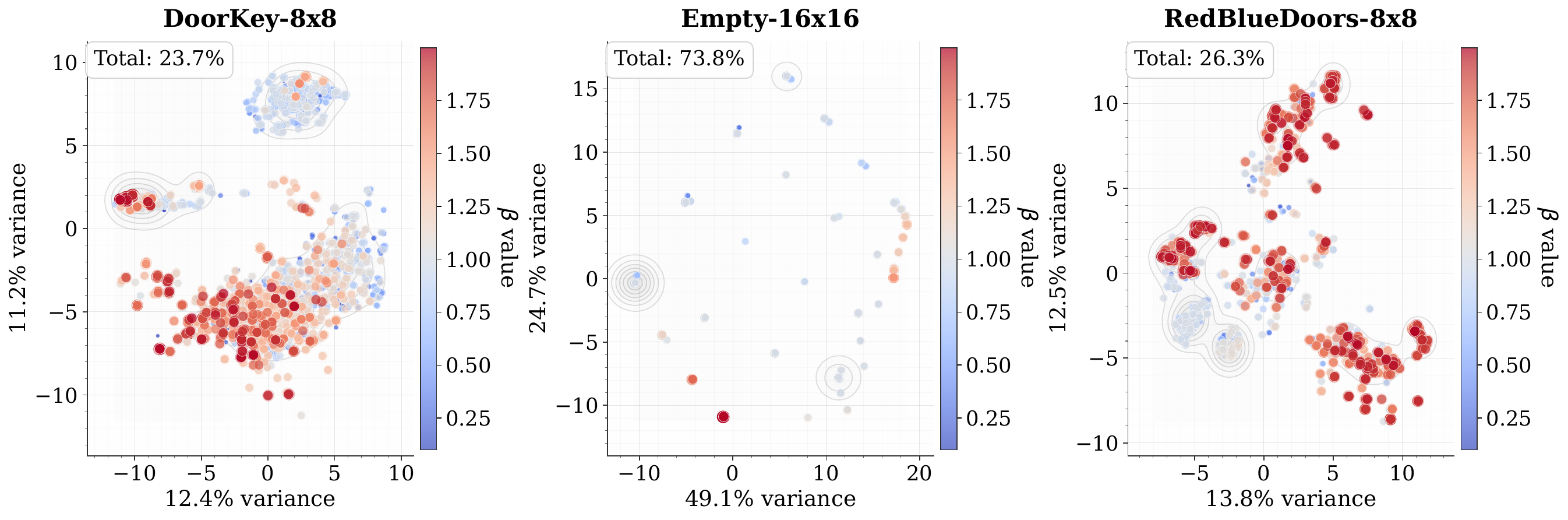}
    \caption{Principal component projections of learned state representations colored by adaptive $\beta$. In \textit{DoorKey-8x8} and \textit{RedBlueDoors-8x8}, $\beta$ aligns with task-relevant regions of the state space. In \textit{Empty-16x16}, $\beta$ shows no systematic relationship with the representation geometry.}
    \label{fig:pca_distribution_stages}
\end{figure*}

Figures~\ref{fig:beta_distribution_stages} and~\ref{fig:pca_distribution_stages} examine the evolution of the learned state-dependent $\beta$ during training and its alignment with the learned state representations.

\paragraph{Dynamics of the $\beta$ distribution}
Figure~\ref{fig:beta_distribution_stages} reveals qualitatively distinct adaptation patterns across environments with different levels of reward structure.

In \textit{DoorKey-8x8} and \textit{RedBlueDoors-8x8}, 
the learned $\beta$ distributions exhibit a characteristic evolution over training. 
During early training, the distributions remain tightly concentrated around their initialization values, 
indicating that the correlation objective has not yet accumulated sufficient signal to drive adaptation. 
As training progresses, the distributions broaden and gradually develop multimodal structure. 
The emergence of multiple modes suggests that the network partitions the state space into regions requiring distinct intrinsic reward scaling. 
Certain state subsets benefit from amplified exploration incentives, whereas others require suppression of intrinsic motivation to avoid interference with extrinsic objectives.
By the final stage of training, both structured environments exhibit a pronounced shift toward lower $\beta$ values, with the distributions collapsing into narrow peaks around $\beta_{\text{min}}$. This systematic downward trend reflects a reduction in the marginal utility of exploration as policies improve and extrinsic rewards become more reliably attainable. The agent learns to progressively down-weight intrinsic motivation in favor of exploitation. The stability of this convergence supports the effectiveness of the correlation-based objective in environments where exploration causally influences future extrinsic returns.
In contrast, \textit{Empty-16x16} exhibits essentially no adaptation throughout training. The $\beta$ distribution remains narrowly concentrated near its initialization value with negligible variance. This behavior reflects the extreme reward sparsity of the environment: in the absence of intermediate structure linking exploration to eventual success, the correlation between intrinsic rewards and future extrinsic returns approaches zero. Consequently, the correlation objective provides no informative gradient, and the regularization term anchors $\beta$ near its prior.
Notably, this failure to adapt constitutes graceful degradation rather than instability. Learning proceeds successfully under an effectively fixed $\beta$, demonstrating that the base algorithm remains robust when correlation learning signals are uninformative. Rather than diverging or oscillating, the system defaults to a conservative constant scaling, validating the role of regularization. This highlights a fundamental limitation: state-dependent adaptation requires sufficient environmental structure to generate meaningful correlations between exploration and outcomes. When such structure is absent, the mechanism correctly refrains from learning spurious adaptations.

\paragraph{Alignment with the state representation}
Figure~\ref{fig:pca_distribution_stages} visualizes the learned state representations projected onto their first two principal components, with points colored according to their associated $\beta$ values. This projection illustrates how adaptive intrinsic weighting aligns with the geometry of the learned state space.

In \textit{DoorKey-8x8}, where the first two components capture 23.7\% of the total variance, $\beta$ values exhibit clear spatial organization. States form coherent clusters with smooth color gradients, indicating that geometrically proximate states are assigned similar intrinsic reward weights. Distinct regions correspond to different $\beta$ regimes: one region concentrates near-neutral values, another exhibits elevated $\beta$, and a third shows suppressed intrinsic weighting. The smooth transitions between these regions indicate that the network generalizes across the representation space rather than memorizing individual states.
\textit{RedBlueDoors-8x8} displays even stronger structure, with 26.3\% of the variance captured by the first two components. The projected state space separates into sharply delineated clusters with distinct $\beta$ assignments.
By contrast, \textit{Empty-16x16} shows no meaningful alignment between $\beta$ and the state geometry, despite the first two components capturing 73.8\% of the variance. $\beta$ values appear randomly distributed across the projection, with nearby states often assigned substantially different values. This absence of structure confirms that the correlation objective does not impose spurious organization when no informative signal is present, instead maintaining near-uniform values with minor fluctuations around initialization.
Across all environments, the PCA visualizations confirm that $\beta$ is learned through the shared encoder rather than through state memorization. Structured environments exhibit coherent spatial organization, whereas uninformative environments exhibit uniformly unstructured assignments.

\begin{figure*}[!h]
    \centering
    \includegraphics[width=\linewidth,height=1.0\textheight,keepaspectratio]{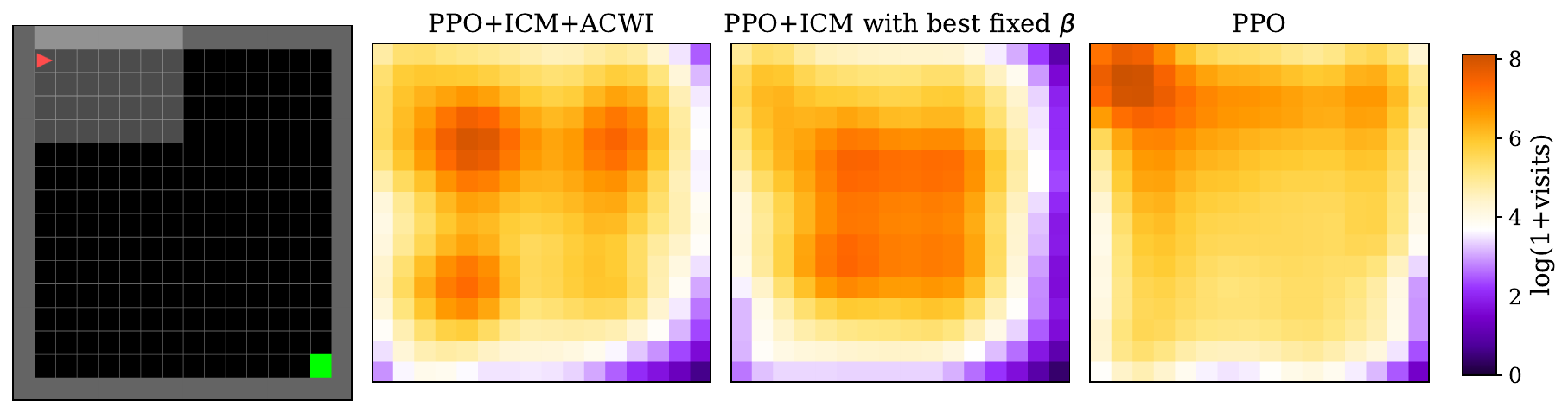}
    \caption{Early training state visitation heatmaps for PPO, PPO+ICM using the best fixed $\beta$, and PPO+ICM+ACWI. Heatmaps are computed during the initial training phase, prior to reliable goal attainment.}
    \label{fig:heatmap_analysis}
\end{figure*}

\paragraph{Exploration behavior}
Figure~\ref{fig:heatmap_analysis} compares early-stage state visitation patterns across methods during the initial phase of training, when extrinsic rewards are sparse or absent. Heatmaps show logarithmic visitation counts.

The PPO baseline exhibits severe exploration failure, with visitation heavily concentrated near the spawn location. PPO+ICM using the best fixed $\beta$ achieves substantially broader coverage. Visitation spans most of the environment, confirming that intrinsic motivation drives systematic exploration. However, coverage is largely uniform across states, indicating a lack of prioritization among regions with differing task relevance.
PPO+ICM+ACWI achieves similarly broad coverage but exhibits subtle structural differences. Certain regions show moderately increased visitation, while others receive reduced emphasis, suggesting the early emergence of state differentiation. Although extrinsic signals remain weak at this stage, preliminary correlation signals begin shaping exploration preferences.
The similarity between the fixed and adaptive approaches during early training is expected. When extrinsic rewards are absent, anchoring $\beta$ near its prior. Nonetheless, the adaptive method implicitly prepares for later adaptation by learning structured representations. Once extrinsic rewards become available, this enables rapid reweighting of intrinsic motivation based on accumulated experience.

\section{Conclusion}
We introduced ACWI (Adaptive Correlation-Weighted Intrinsic), a practical method for learning state-dependent intrinsic-reward scaling via a lightweight Beta network trained with a correlation-based correlation objective. ACWI aligns scaled intrinsic bonuses with discounted future extrinsic returns without costly unrolled meta-gradients, integrating seamlessly with standard policy optimizers. Empirically, ACWI improves sample efficiency and stabilizes learning across sparse-reward MiniGrid tasks with negligible overhead. Analysis confirms that ACWI focuses exploration on task-relevant states when extrinsic signals are informative, gracefully degrading to fixed scaling under extreme sparsity. Future work includes extending to other intrinsic modules, multi-task settings, and developing theoretical guarantees for the correlation objective.




\begin{thebibliography}{00}

\bibitem{mnih2015}
V. Mnih et al., 
``Human-level control through deep reinforcement learning,'' 
\textit{Nature}, vol. 518, no. 7540, pp. 529--533, 2015.

\bibitem{silver2016}
D. Silver et al., 
``Mastering the game of Go with deep neural networks and tree search,'' 
\textit{Nature}, vol. 529, pp. 484--489, 2016.

\bibitem{francois2018}
V. François-Lavet, P. Henderson, R. Islam, M. G. Bellemare, and J. Pineau,
``An introduction to deep reinforcement learning,''
\textit{Foundations and Trends in Machine Learning}, vol. 11, no. 3--4, pp. 219--354, 2018.

\bibitem{sutton2018}
R. S. Sutton and A. G. Barto, 
\textit{Reinforcement Learning: An Introduction}, 2nd ed., Cambridge, MA: MIT Press, 2018.

\bibitem{bellemare2016}
M. G. Bellemare, S. Srinivasan, G. Ostrovski, T. Schaul, D. Saxton, and R. Munos, 
``Unifying count-based exploration and intrinsic motivation,'' 
in \textit{Advances in Neural Information Processing Systems (NeurIPS)}, vol. 29, pp. 1471--1479, 2016.

\bibitem{tang2017}
H. Tang et al., 
``\#Exploration: A study of count-based exploration for deep reinforcement learning,'' 
in \textit{Advances in Neural Information Processing Systems (NeurIPS)}, vol. 30, pp. 2753--2762, 2017.

\bibitem{pathak2017}
D. Pathak, P. Agrawal, A. A. Efros, and T. Darrell, 
``Curiosity-driven exploration by self-supervised prediction,'' 
in \textit{Proc. 34th Int. Conf. on Machine Learning (ICML)}, vol. 70, pp. 2778--2787, 2017.

\bibitem{burda2018_rnd}
Y. Burda, H. Edwards, A. Storkey, and O. Klimov, 
``Exploration by random network distillation,'' 
in \textit{Proc. Int. Conf. on Learning Representations (ICLR)}, 2019.


\bibitem{raileanu2020}
R. Raileanu and T. Rockt{\"a}schel, 
``RIDE: Rewarding impact-driven exploration for procedurally-generated environments,'' 
in \textit{Proc. Int. Conf. on Learning Representations (ICLR)}, 2020.

\bibitem{ostrovski2017}
G. Ostrovski, M. G. Bellemare, A. van den Oord, and R. Munos,
``Count-based exploration with neural density models,''
in \textit{Proc. 34th Int. Conf. on Machine Learning (ICML)}, vol. 70, pp. 2721--2730, 2017.

\bibitem{chen2022eipo}
E. Chen, Z.-W. Hong, J. Pajarinen, and P. Agrawal, 
``Redeeming intrinsic rewards via constrained optimization,'' 
in \textit{Advances in Neural Information Processing Systems (NeurIPS)}, vol. 35, pp. 24041--24054, 2022.

\bibitem{yuan2023airs}
M. Yuan, B. Li, X. Jin, and W. Zeng, 
``Automatic intrinsic reward shaping for exploration in deep reinforcement learning,'' 
in \textit{Proc. 40th Int. Conf. on Machine Learning (ICML)}, vol. 202, pp. 40356--40375, 2023.

\bibitem{schulman2017ppo}
J. Schulman, F. Wolski, P. Dhariwal, A. Radford, and O. Klimov, 
``Proximal policy optimization algorithms,'' 
\textit{arXiv preprint arXiv:1707.06347}, 2017.

\bibitem{oudeyer2007}
P.-Y. Oudeyer, F. Kaplan, and V. Hafner,
``Intrinsic motivation systems for autonomous mental development,''
\textit{IEEE Transactions on Evolutionary Computation}, vol. 11, no. 2, pp. 265--286, 2007.

\bibitem{schmidhuber2010}
J. Schmidhuber,
``Formal theory of creativity, fun, and intrinsic motivation,''
\textit{IEEE Transactions on Autonomous Mental Development}, vol. 2, no. 3, pp. 230--247, 2010.

\bibitem{strehl2008}
A. L. Strehl and M. L. Littman,
``An analysis of model-based interval estimation for Markov decision processes,''
\textit{Journal of Computer and System Sciences}, vol. 74, no. 8, pp. 1309--1331, 2008.

\bibitem{taiga2020}
A. A. Taiga et al., 
``On bonus based exploration methods in the Arcade Learning Environment,'' 
in \textit{Proc. Int. Conf. on Learning Representations (ICLR)}, 2020.

\bibitem{badia2020_agent57}
A. P. Badia et al., 
``Agent57: Outperforming the Atari human benchmark,'' 
in \textit{Proc. 37th Int. Conf. on Machine Learning (ICML)}, vol. 119, pp. 507--517, 2020.

\bibitem{badia2020_ngu}
A. P. Badia et al., 
``Never Give Up: Learning directed exploration strategies,'' 
in \textit{Proc. Int. Conf. on Learning Representations (ICLR)}, 2020.

\bibitem{ng1999}
A. Y. Ng, D. Harada, and S. J. Russell, 
``Policy invariance under reward transformations: Theory and application to reward shaping,'' 
in \textit{Proc. 16th Int. Conf. on Machine Learning (ICML)}, pp. 278--287, 1999.

\bibitem{houthooft2016}
R. Houthooft, X. Chen, Y. Duan, J. Schulman, F. De Turck, and P. Abbeel,
``VIME: Variational information maximizing exploration,''
in \textit{Advances in Neural Information Processing Systems (NeurIPS)}, vol. 29, pp. 1109--1117, 2016.

\bibitem{kim2019}
H. Kim, J. Kim, Y. Jeong, S. Levine, and H. O. Song,
``EMI: Exploration with mutual information,''
in \textit{Proc. 36th Int. Conf. on Machine Learning (ICML)}, vol. 97, pp. 3360--3369, 2019.

\bibitem{jaderberg2017}
M. Jaderberg et al., 
``Reinforcement learning with unsupervised auxiliary tasks,'' 
in \textit{Proc. Int. Conf. on Learning Representations (ICLR)}, 2017.

\bibitem{mirowski2017}
P. Mirowski et al., 
``Learning to navigate in complex environments,'' 
in \textit{Proc. Int. Conf. on Learning Representations (ICLR)}, 2017.

\bibitem{zahavy2020_stac}
T. Zahavy et al., 
``Self-tuning deep reinforcement learning,'' 
in \textit{Advances in Neural Information Processing Systems (NeurIPS)}, vol. 33, pp. 11839--11850, 2020.

\bibitem{schulman2015gae}
J. Schulman, P. Moritz, S. Levine, M. Jordan, and P. Abbeel, 
``High-dimensional continuous control using generalized advantage estimation,'' 
in \textit{Proc. Int. Conf. on Learning Representations (ICLR)}, 2016.

\bibitem{finn2017}
C. Finn, P. Abbeel, and S. Levine, 
``Model-agnostic meta-learning for fast adaptation of deep networks,'' 
in \textit{Proc. 34th Int. Conf. on Machine Learning (ICML)}, vol. 70, pp. 1126--1135, 2017.

\bibitem{pascanu2013}
R. Pascanu, T. Mikolov, and Y. Bengio, 
``On the difficulty of training recurrent neural networks,'' 
in \textit{Proc. 30th Int. Conf. on Machine Learning (ICML)}, pp. 1310--1318, 2013.

\bibitem{chevalier2018minigrid}
M. Chevalier-Boisvert, L. Willems, and S. Pal,
``Minimalistic gridworld environment for openai gym,''
in \textit{Proc. ICLR Workshop on Representation Learning for RL}, 2018.
\end{thebibliography}
\end{document}